\documentclass[conference]{IEEEtran}
\IEEEoverridecommandlockouts
\usepackage[utf8]{inputenc}
\usepackage[english]{babel}
\usepackage{amsmath,amssymb,amsfonts}
\usepackage{algorithmic}
\usepackage{graphicx}
\usepackage{textcomp}
\usepackage{xcolor}
\usepackage{csquotes}

\usepackage[backend=biber, sorting=none, style=ieee]{biblatex}
\begin{document}

\title{Curiosity Based Reinforcement Learning on Robot Manufacturing Cell}

\author{
\IEEEauthorblockN{Mohammed Sharafath Abdul Hameed, Md Muzahid Khan, Andreas Schwung}

\IEEEauthorblockA{Department of Automation Technology\\
South Westphalia University of Applied Sciences\\
Soest, Germany.\\
\emph{sharafath.mohammed, khan.mdmuzahid,  schwung.andreas@fh-swf.de}}\\
}
\maketitle

\begin{abstract}
This paper introduces a novel combination of scheduling control on a flexible robot manufacturing cell with curiosity based reinforcement learning. Reinforcement learning has proved to be highly successful in solving tasks like robotics and scheduling. But this requires hand tuning of rewards in problem domains like robotics and scheduling even where the solution is not obvious. To this end, we apply a curiosity based reinforcement learning, using intrinsic motivation as a form of reward, on a flexible robot manufacturing cell to alleviate this problem. Further, the learning agents are embedded into the transportation robots to enable a generalized learning solution that can be applied to a variety of environments. In the first approach, the curiosity based reinforcement learning is applied to a simple structured robot manufacturing cell. And in the second approach, the same algorithm is applied to a graph structured robot manufacturing cell. Results from the experiments show that the agents are able to solve both the environments with the ability to transfer the curiosity module directly from one environment to another. We conclude that curiosity based learning on scheduling tasks provide a viable alternative to the reward shaped reinforcement learning traditionally used.
\end{abstract}


\section{Introduction}
Reinforcement Learning (RL) is becoming a popular algorithm to solve complex games and tasks like Atari~\cite{mnih2015human}, physics simulation~\cite{Todorov.2012}, Go~\cite{Silver.2018} and Chess~\cite{schrittwieser2019mastering}, Robotics~\cite{Gu.2016}, and optimization problems~\cite{Cunha.2020} etc. In these approaches, the RL agent is given a continuous supply of dense extrinsic reward by the environment for the action it takes. While providing a reward or defining a reward function is not a problem in common gaming tasks, it quickly becomes cumbersome in complex engineering tasks like the optimization problem~\cite{Xia.2020}. And traditionally when the rewards are not directly available, they need to be shaped to guide the agent in the direction of an optimal solution. This approach creates three problems: one, this introduces a huge inductive bias in the reward scheme that it cripples the generalized learning of the RL agent from the start. So much so that it might be only able to solve in the known solution space; two, reward shaping is a notoriously difficult engineering problem, where un-optimized rewards could break the learning process or provide only sub-optimal solutions; and finally, it creates solutions which cannot be transferred to new environments, since the rewards have been heavily tuned for a particular environment. In this paper, we propose to apply curiosity based learning to a flexible Robot Manufacturing Cell.

In~\cite{schwung2017}, a Robot Manufacturing Cell (RMC) was proposed as an optimization problem similar to the scheduling problems encountered in the production planning. And in~\cite{schwung2019}, the same RMC was solved using central and distributed advantage actor critic (A2C), through hand engineered reward signals. Similar reward shaping methods are also followed in~\cite{Park.2020}~\cite{Gabel.2008}~\cite{Waschneck.2018}, where rewards for each of the scheduling problem is hand engineered to suit the environment they work in. This obviously creates a problem of non-generalized solutions, where the learning of an agent from one environment is not transferable to an another, even though they both solve similar optimization problems.

To overcome this limitation of the reinforcement learning application in engineering problems, in this paper we apply curiosity~\cite{Pathak.2017} to the RMC environment. Curiosity uses the intrinsic motivation of the agent to create rewards based on the surprise factor of the agents familiarity with the environment states. Through the application of curiosity in reinforcement learning of the RMC we overcome all the problems stated above. The main contributions of this paper as follows:

\begin{itemize}
    \item We propose curiosity as an effective tool to eliminate the inductive bias induced by the reward shaping normally used in reinforcement learning of scheduling tasks. Further, curiosity eliminates complex reward shaping for engineering tasks like scheduling.
    \item We show that a direct transfer of the reward function from one manufacturing environment to another is possible without any tuning requirements.
    \item We combine curiosity with gradient monitoring, an effective approach to stabilize training. We apply the curiosity-based RL to two multi-robot environments where we particularly show the improvements achieved by the proposed approach in both environments.
\end{itemize}

\par The paper is organized as follows. Section~\ref{section:RMC} introduces the learning problem, RMC, in detail, describing the operations and setup. Section~\ref{section:RL} introduces the RL algorithms used in this paper along with further optimizations that were used, while Section~\ref{section:icm}, Section~\ref{section:GM}, and Section~\ref{section:CL} explain the curiosity module along with other optimizations used. Section~\ref{section:results} describes the results that were achieved on the RMC through the application of the RL along with its optimizations, while Section~\ref{section:conclusion} provides the conclusion with future scope of work.

\section{Robot Manufacturing Cell}
\label{section:RMC}
\begin{figure}[htp]
    \centering
    \includegraphics[width=8cm]{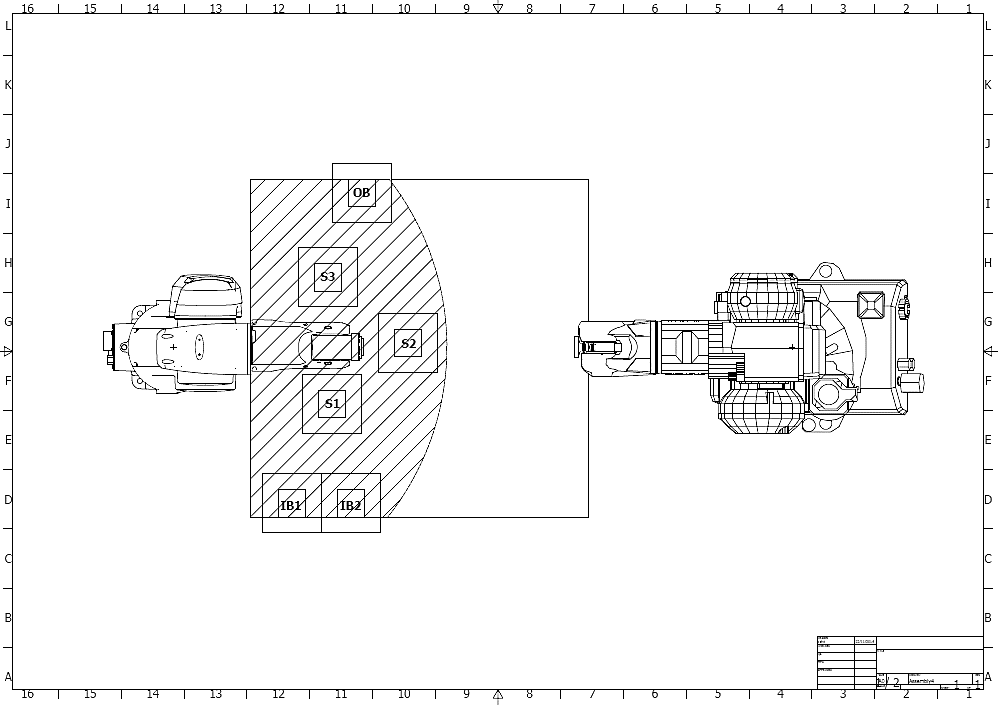}
    \caption{Schematic drawing of the robot manufacturing cell}
    \label{fig:rmc}
\end{figure}

In this paper, RL model is applied to a cooperative self-learning RMC~\cite{schwung2017}. The RMC consists of two industrial robots, an Adept Cobra i600 SCARA-robot and an ABB  IRB1400 6-DOF-robot, both of which share a common work platform as shown in Figure~\ref{fig:rmc}. The setup has six stations totally, two input stations (IB1, IB2), three processing or work stations (M1, M2, M3) and one output station (OB). There are two different types of work-piece in this system (WP-1, WP-2). The WP-1 goes through IB1, M1, M2, M3, and then OB, while WP-2 goes through IB2, M2, M1, M3 and then OB. Both the robots have their own proprietary control software. Hence a supervisory Siemens Programmable Logic Control (PLC) system is developed to control and coordinate the robot movement. Two experiments are conducted to demonstrate the applicability of the curiosity based approach on scheduling problems. The first experiment is conducted on a simple structured Robot Manufacturing Cell (sRMC) environment, similar to~\cite{schwung2019} and the second experiment is conducted on a graph-network based Robot Manufacturing Cell (gRMC) environment, which was developed in~\cite{hameed2020graphs}. gRMC is a scalable variant of the sRMC environment, where the machines and the buffers are represented by the nodes of a graph while the connections between them are expressed as edges. A work-piece is transported between the nodes when the edge is activated.\par

\begin{figure}[htp]
    \centering
    \includegraphics[width=8cm]{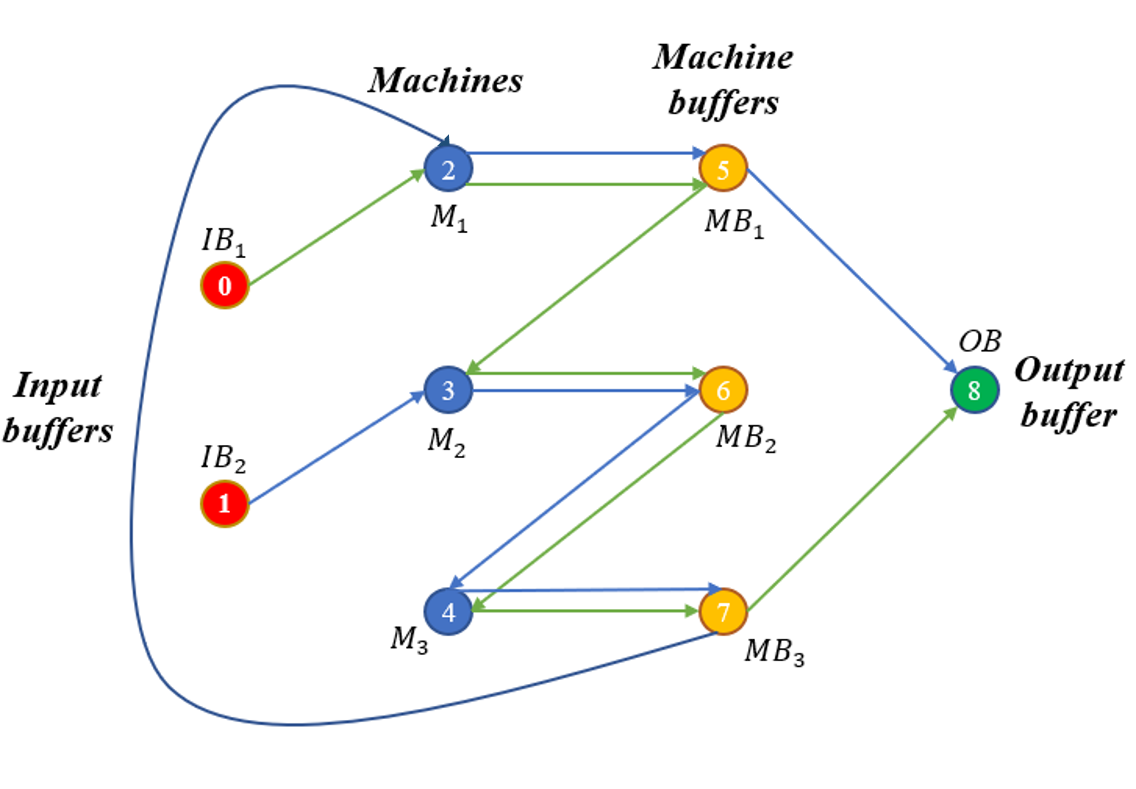}
    \caption{In the gRMC environment, the nodes are indexed from 0-8 and are colored according to their function, while the different colored lines represent the two work-pieces to be processed and transported.}
    \label{fig:gRMC}
\end{figure}

In sRMC no buffers are considered for the processing stations  environment, i.e., there is no queue in front or after the processing stations. The RL agent's actions here are the movements of the work-pieces between the processing stations. While in gRMC three buffer-stations are considered, one after each of the processing stations (MB1, MB2, MB3) as shown in Figure~\ref{fig:gRMC}. Here the agent's action is the activation of the edges connecting the nodes. If a node is activated then the work-piece is moved from the ’from’-node to the ’to’-node. Two different types of edges are defined in the system, to move the different work-pieces. Both the work-pieces are embedded with the RFID chip which gives the required information to the robots.

\section{Reinforcement Learning}
\label{section:RL}
RL is a field of machine learning where an agent is trained to take a sequence of decisions on an environment, modeled as a Markov Decision Process (MDP), to maximize the cumulative reward signal. The RL system has two major components namely, the agent and the environment. The agent interacts with the environment, takes an action $a_t$ based on the agent's policy $\pi$($a_t|s_t)$ from its current state $s_t$ and moves to a new state $s_{t+1}$. Based on the action taken by the agent, the environment sends a reward signal $r_t$ to the agent. The goal of the agent is to find an optimal policy that maximizes the cumulative reward~\cite{sutton2018reinforcement}. Hence, the agent tries to maximize the expected return $\mathbb{E}[R_t|\pi]$ with the discounted reward given by $R$ = $\sum_t\gamma^t r_t$ and discount factor 0 $\leq\gamma\leq$ 1. The MDP is defined by the tuple ($\mathcal{S,A,P,R}, p_0$), with the set of states $\mathcal{S}$, the finite set of actions $\mathcal{A}$, a transition model $\mathcal{P}$ for a given state $s$ and an action $a$, the reward function $\mathcal{R:S\times A \times S \rightarrow \mathbb{R}}$, which provides a reward $r$ for each state transition $s_t\rightarrow s_{t+1}$ with an initialization probability $p_0$. \par

The policy is defined using either value based methods or policy gradient methods. Value based methods use either a  \textit{state-value function}, $v(s) = \mathbb{E}[R_{t}|S_{t}=s]$ or a \textit{state-action-value function}, $q(s, a) = \mathbb{E}[R_{t}|S_{t}=s, A_{t} = a]$. The policy is then defined by an $\epsilon-greedy$ strategy where greedy actions are given by $\pi(a|s) = argmax(q(s, \mathcal{A}))$, where the agent exploits, or a non-greedy action is given, where the agent explores. In policy gradient methods, a \textit{parameterized policy}, $\pi(a|s,\theta)$, is used to take action. The policy can still be improved using value functions as seen in~\cite{mnih2015human}. The the agent's objective is to derive an optimal policy, $\pi^*(a|s),$ that maximizes the reward collection. Correspondingly, the policy is updated in a way to maximize a cost function $J(\theta_{t})$ given by,

\begin{equation}
    \label{eq:policy update}
    \theta_{t+1} = \theta_{t} + \rho \nabla J(\theta_{t}),
\end{equation}

where $\theta$ is the policy parameters of $\pi$ and $\rho$ is the learning rate.

\subsection{Advantage Actor Critic (A2C):} Advantage Actor Critic is a policy gradient algorithm which combines policy gradient method with value based approach like SARSA~\cite{sutton2018reinforcement}. In A2C, the actor tries to find the optimal policy and the critic evaluates the actor's policy. Here, the actor refers to the policy $\pi(a|s,\theta_1)$ and the critic refers to the value function $\nu (s, \theta_2)$, where $\theta_1$ is the parameter of the actor and the $\theta_2$  is the parameter of the critic. The cost function equation of the A2C is given by Eqn.~\ref{eq:a2c_cost} and~\ref{eq:a2c_adv} respectively. \par

\begin{equation}
\label{eq:a2c_cost}
    J(\theta)=E_{\sim\pi_\theta} \bigg[\sum\limits_{(s_t,a_t)\epsilon}log\pi_\theta(a_t,s_t) A_{\pi_\theta}\bigg]
\end{equation}

\begin{equation}
\label{eq:a2c_adv}
    A_{\pi_\theta}(s_t,a_t)=Q_{\pi_\theta}(s_t,a_t)-V_{\pi_\theta}(s_t)\\
\end{equation}

\section{Curiosity Driven Learning}
\label{section:icm}
\begin{figure*}[htp]
    \centering
    \includegraphics[width=10cm, height=5cm]{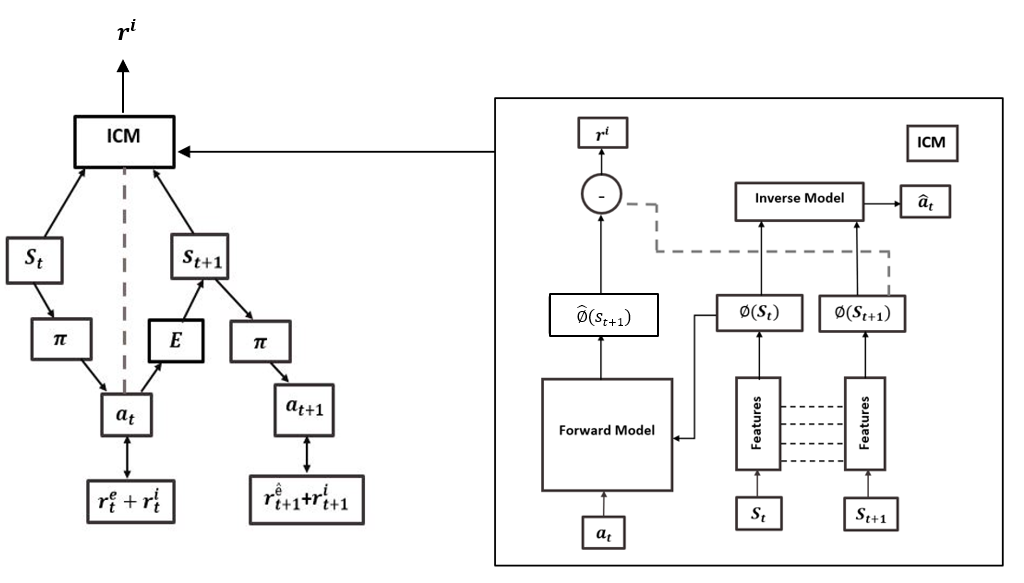}
    \caption{Graphical representation of the Intrinsic Curiosity Module}
    \label{fig:ICM}
\end{figure*}

Children often learn through intrinsic motivation such as curiosity. Psychologists define a behavior as intrinsically motivated when a human works for his own sake. Similarly, intrinsic motivation in RL keeps an agent engaged in the exploration of new knowledge and skills when the rewards are rare~\cite{chentanez2005intrinsically}. There are lots of approaches to develop the intrinsic reward of an agent like prediction error, prediction uncertainty, or development of the forward model trained along with the policy of an agent~\cite{burda2018large}. These approaches drive the agent to explore the environment in the absence of a dense extrinsic reward. In this paper, we use prediction error as the intrinsic reward, which is produced by the agent itself with a model called Intrinsic Curiosity Model (ICM). ICM is composed of two subsystems: an inverse dynamic model that produce an intrinsic reward signal and forward dynamic model that outputs a sequence of actions which guides the agent to maximize the reward signal~\cite{Pathak.2017}. Figure~\ref{fig:ICM} represents the schematic architecture of the ICM. The ICM encodes the current state $s_t$ and the next state $s_{t+1}$ into the feature of current state $\phi(s_t)$ and feature of next state $\phi(s_t+1)$. The feature of current state $\phi(s_t)$ and the feature of next state $\phi(s_t+1)$ are trained with inverse dynamic model that predicts action $\hat{a}_t$~\cite{Pathak.2017}. Mathematically, this transition can be represented as,

\begin{equation}
    \hat{a}_t= g(s_t,s_{t+1};\theta_I)
\end{equation}

where $g$ is a soft-max distribution function of all possible actions that could be taken and $\theta_I$ is the neural network parameter. The loss function of the inverse dynamic model can represent by the following,

\begin{equation}
    {\min_{\theta_{I}}}= L_I(\hat{a_t},a_t)
\end{equation}

where $L_I$ the loss function that calculates the deviation between the actual action $a_t$ and the predicted action $\hat{a_t}$. The forward dynamic model takes $\phi({s_t})$ and action $a_t$ as the inputs and outputs the predicted feature representation $\hat\phi(a_{t+1})$ of next state $s_{t+1}$. The forward dynamic model can be represented by the following,

\begin{equation}
    \hat\phi(s_{t+1})=f(\phi(s_t),a_t;\theta_F)
\end{equation}

where $f$ is the learned function and $\theta_F$ is the ICM neural network parameter. The loss function of the forward model is given below:

\begin{equation}
    L_F(\phi(s_t),\hat\phi(s_{t+1})=\frac{1}{2}||\phi(s_{t+1})-\phi(s_t)||^{2}_2
\end{equation}

where $L_F$ is the loss function that calculates the variance between the predicted feature representation of next state $\hat\phi(a_{t+1})$ and the actual predicted feature of next state $\phi(s_t+1)$. So, the intrinsic reward $r_t^i$ can be calculated as: 

\begin{equation}
     r_t^i=\frac{\eta}{2}||\phi(s_{t+1})-\phi(s_t)||^{2}_2
\end{equation}

where $\eta>0$ is the scaling factor of the intrinsic reward signal. The intrinsic reward is generated by the agent at time $t$. Besides intrinsic rewards, the agent may also get extrinsic reward $r_t^e$ from the environment.

To generate the intrinsic reward signal and to optimize the ICM, the inverse dynamic model and forward dynamic model loss should be combined. So, the overall optimization problem of ICM that helps the agent to learn can be represented as:
\begin{equation}
      \min_{\theta_P,\theta_I,\theta_F}\Bigg[{-\lambda E_{\pi(s_t;\theta_P)}}\bigg[\sum\nolimits_{t} r_t\bigg]+(1-\beta)L_I+\beta L_F\Bigg]
\end{equation}
where 0$\leq$$\beta$$\leq$1 is a scaling factor that weights the inverse model against the forward model, and ${\lambda>0}$ is a scaling factor of policy gradient loss against the intrinsic reward~\cite{Pathak.2017}. In this paper, to speed up the learning process of the RL agents two optimization techniques called Gradient Monitoring and Curriculum Learning are used. They are explained further below.

\subsection{Gradient Monitoring}
\label{section:GM}
RL has a non stationary learning problem. This makes ensuring stability in learning a critical part of RL. In this paper an optimization method called Gradient Monitoring (GM) is used to stabilize the RL training of the RMC environments. GM provides for a faster convergence and better generalization performance of the trained RL agent, for a small computational overhead, by steering the gradients during the learning process towards the most important weights. The added benefit of using GM is that it allows for a continuous adjustment of the neural network's utilization capacity, i.e., the size of the neural network need not be tuned with as much care since GM automatically derives the required size of the neural network required for the training process~\cite{hameed2020gradient}.

\subsection{Curriculum Learning}
\label{section:CL}
Humans learn things efficiently, when the tasks are organized in such a way that the complexity levels increase gradually. This strategy is called Curriculum Learning (CL). CL is a training process where the criterion of the training process is co-related with different set of weights. It is a form of re-weighting of the training distribution. At the beginning, the associated weights represent the simpler task, that can be learned most easily. The next training criterion involves a small change in the weighting of task that increases the probability of sampling slightly more complex task. At the end, the re-weighting of the tasks is uniform, and the final target is trained~\cite{bengio2009curriculum}. In RL, sometimes it becomes difficult for the RL agent to handle complex tasks directly and to solve this problem CL technique is used. \par

\section{Experimental Results}
\label{section:results}
In this paper, two experiments are conducted. First sRMC is solved using ICM and CL, since the final task of planning of 20 WP-1 and 20 WP-2 was too difficult for the agent. Then in the same experiment, GM is introduced which helps the agent to converge without the use of CL. Next in the second experiment, the ICM used in sRMC is used in the gRMC environment. Here, only CL is used since the overhead GM did not provide a corresponding gain. The results thence obtained are discussed below in two parts based on their environment, sRMC and gRMC. The parts output in each of the case is reported, along with the rewards collected by in setup. \par

\subsection{sRMC:} 
\begin{figure}[t]
    \centering
    \includegraphics[width=8cm]{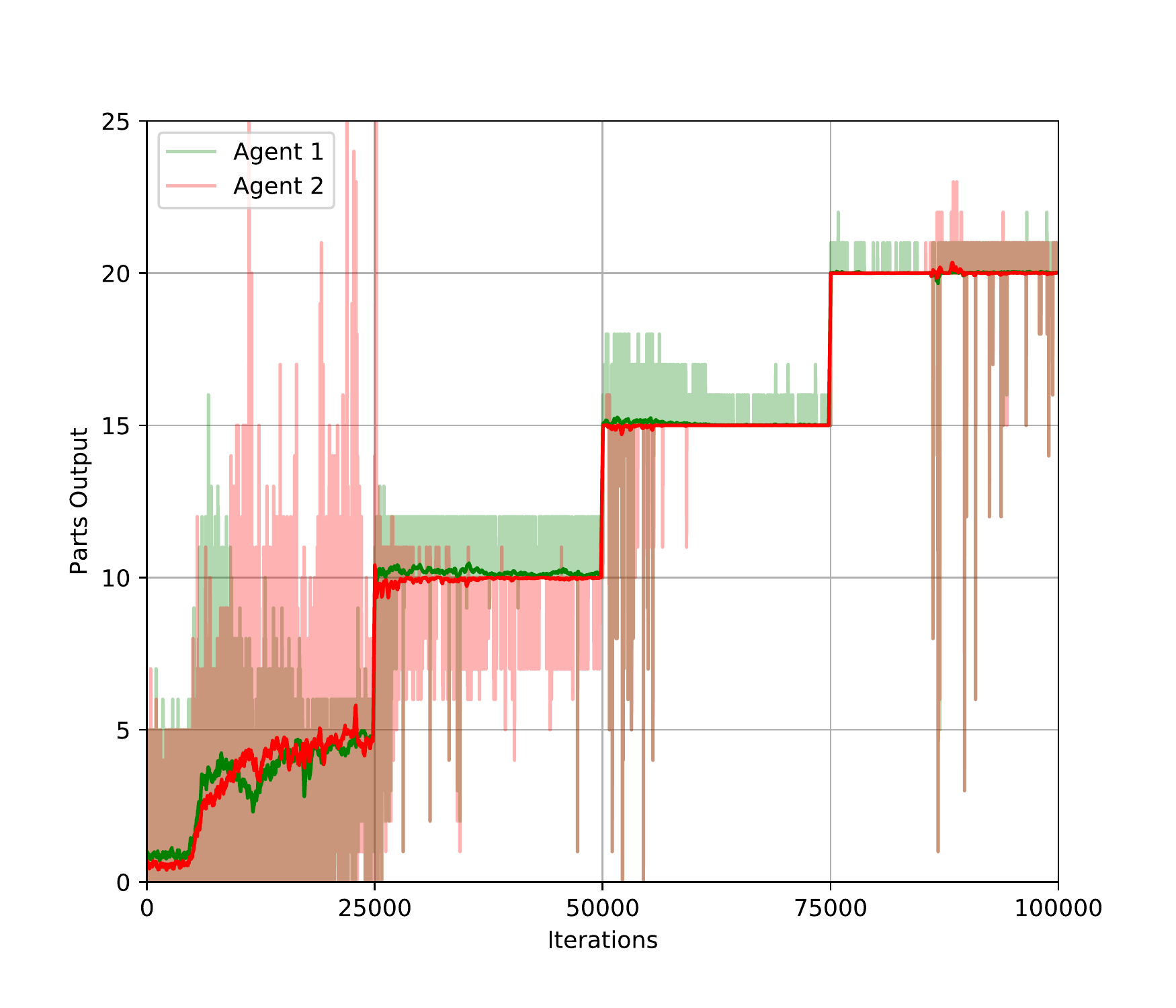}
    \caption{Parts Output (sRMC without gradient monitoring)}
    \label{fig:output_parts}
\end{figure}

In this section, first the results of curriculum learning without gradient monitoring approach in sRMC environment is discussed. The result in parts output, WP-1 and WP-2, with respect to iterations is shown in Figure~\ref{fig:output_parts}. Here, the target is 20 work-pieces for each agent. Initially, without the additional optimizations the agents were able to solve only seven work-pieces each as opposed to the target of 20 each. Hence, CL was introduced to divide the final target into smaller targets of increasing difficulty, i.e. target of 5, 10, 15, and 20 work-pieces of WP-1 and WP-2. Both the agents are trained for $1e^5$ episodes and the convergence for 20 target is achieved after $0.78e^5$ episodes.

\begin{figure}[t]
    \centering
    \includegraphics[width=\linewidth]{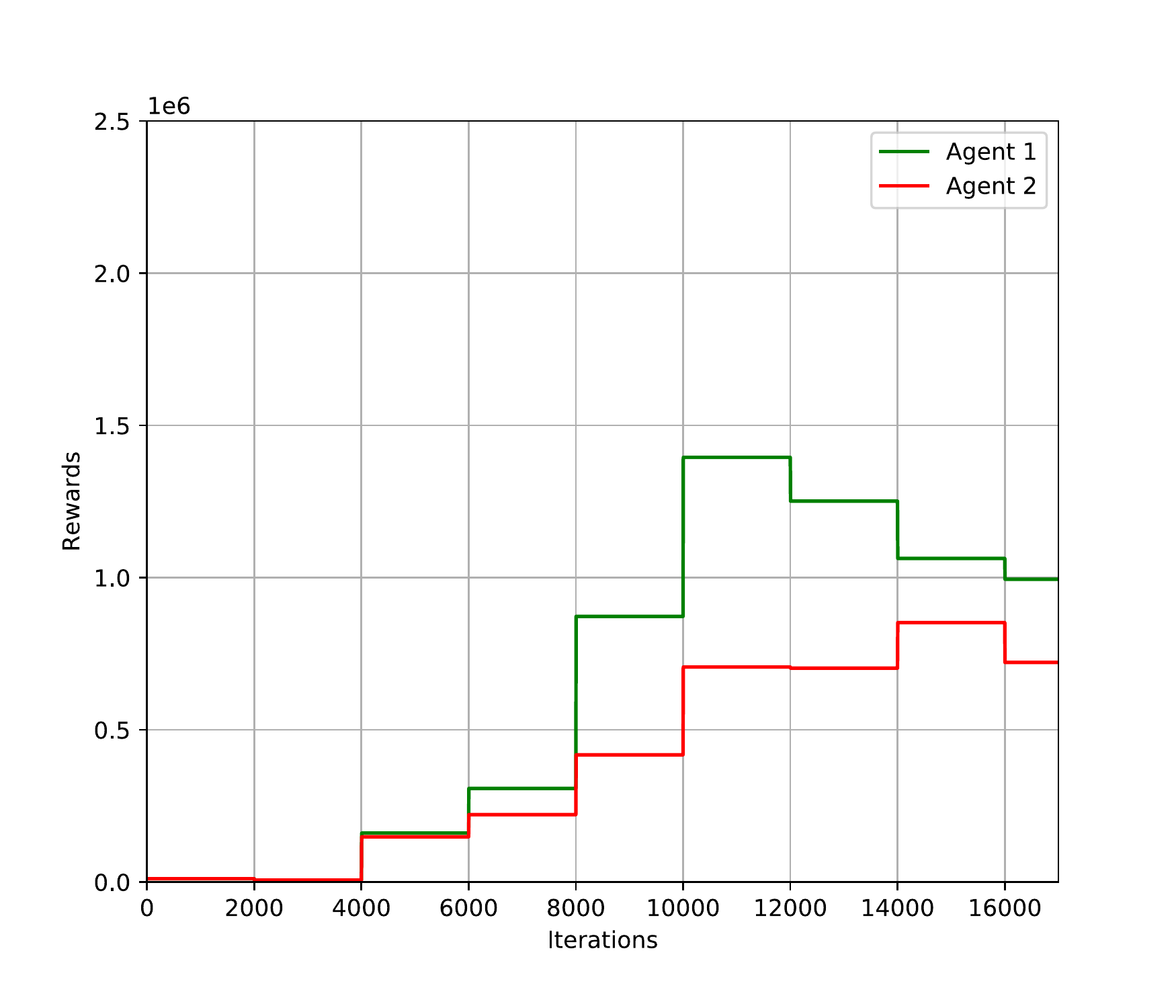}
    \caption{Combined Rewards (sRMC with gradient monitoring)}
    \label{fig:Rewards_gm}
\end{figure}

To increase the convergence speed of the agents, GM was introduced into the agent learning the sRMC environment. The improvement posed by GM was good enough to remove the CL completely from the training regime of sRMC environment. This quickened the learning progress as shown in Figure~\ref{fig:output_parts}. A target of 20 work-pieces for each agent was set directly rather than dividing the targets into smaller tasks. The agents solved the target directly in $5e^3$ steps compared to the $0.78e^5$ without GM. With this, the objective of training the agents on the sRMC environment was completed. The results graph shows an interesting trend. Figure~\ref{fig:Rewards_gm} shows the combined, intrinsic and extrinsic, rewards collected by the agents. The rewards graph can be used to describe the agent's learning behavior. Normally, in extrinsic reward-based RL algorithm, the slope of the reward curves becomes zero after reaching a solution. But the intrinsic reward-based system works differently. When the agent remains curious about completing the target, it generates the intrinsic reward by itself. But once the agent finishes exploring its environment and finds the an optimal policy, the agent remains no more curious and loses its motivation, hence receives less reward for the same task. That means, intrinsic rewards obviously reduce once the agent has converged. This is illustrated in the Figure~\ref{fig:Rewards_gm}, where after the iteration number $1e5$, the rewards collected reduces. The next objective is to transfer the ICM incorporated here in to another environment without any architectural changes.

\begin{figure}[t]
    \centering
    \includegraphics[width=9cm]{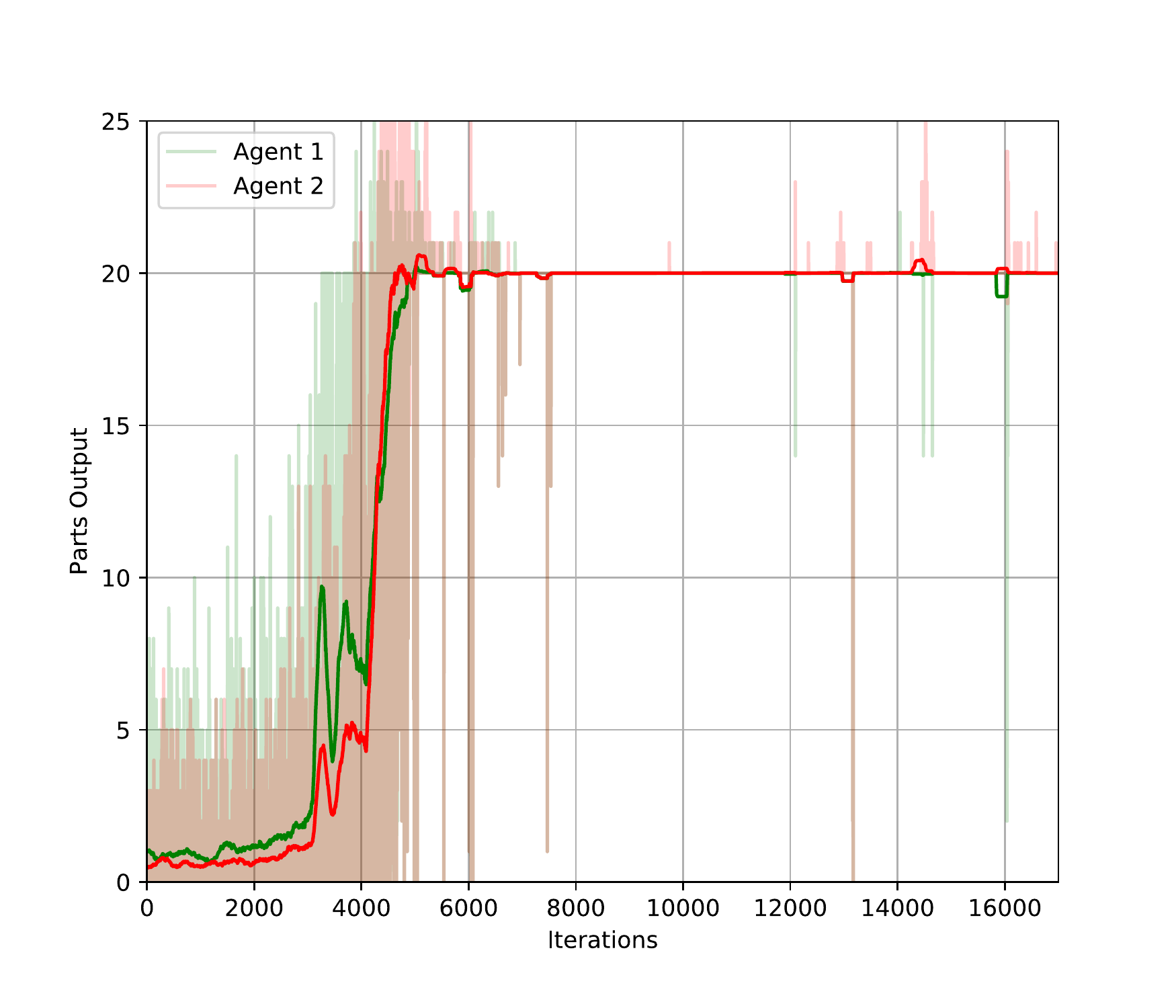}
    \caption{Parts Output (sRMC with gradient monitoring)}
    \label{fig:output_parts_gm}
\end{figure}

\subsection{Graph network based curriculum learning:}
\begin{figure}[t]
\label{fig:output_gRMC}
    \centering
    \includegraphics[width=\linewidth]{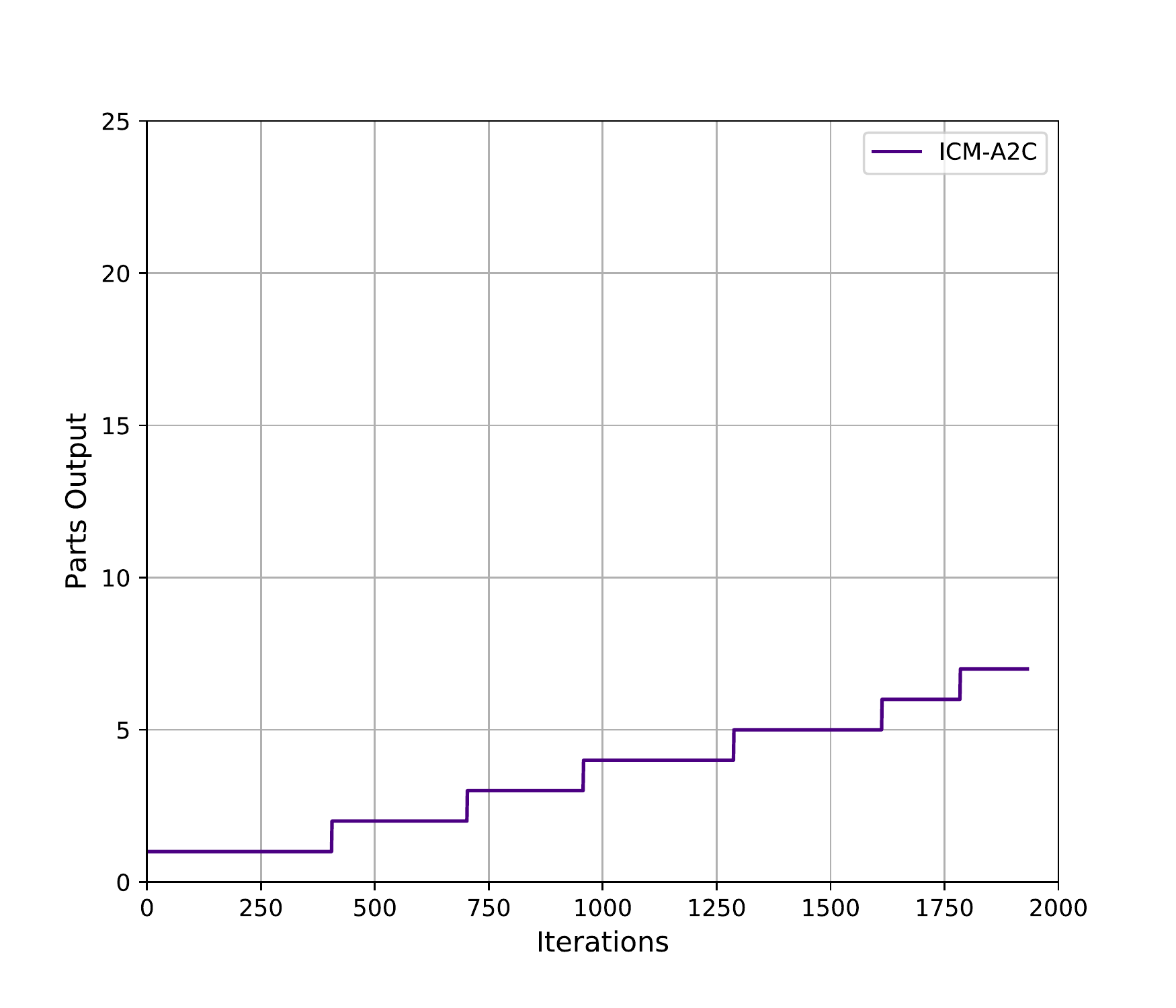}
    \caption{Parts Output (gRMC without gradient monitoring)}
    \label{fig:Int_rew_gm}
\end{figure}

\begin{figure}[htp]
\label{fig:rew_gRMC}
    \centering
    \includegraphics[width=\linewidth]{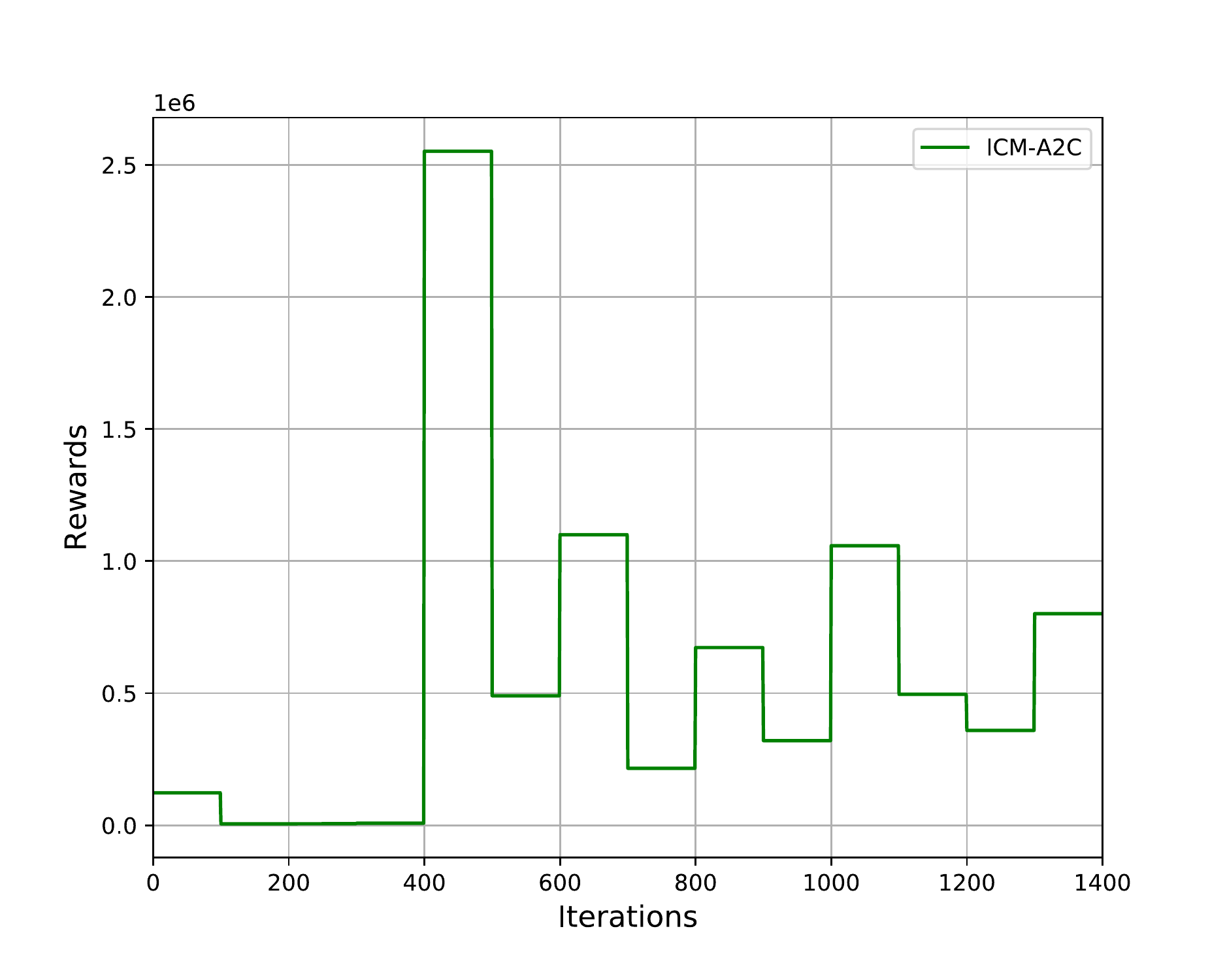}
    \caption{Combined Rewards (gRMC without gradient monitoring)}
    \label{fig:Combined Rewards}
\end{figure}

In the next experiment, the intrinsic curiosity module is plugged to train the agents on gRMC. The learning process was a lot more difficult in this setup. This difficulty was also noted in the reward shaping method, where the agent did not converge in regular training. Hence, first CL was introduced in to the training regime.  The agent's target in curriculum learning was designed keeping in mind the difficulty i.e., targets of 1, 2, 3, 4, etc., were used. The agent solved the targets until a target of seven work-pieces each as shown in Figure~\ref{fig:output_gRMC}, although the training times were extremely high. Another optimization trial using GM produced similar results, but instead increased the training times marginally. This could be because of the complex nature of the gRMC interaction and the higher size of the action space. When looking at Figure~\ref{fig:rew_gRMC} though, the agent starts exploring the environment from its early stage and the agent reaches to the peak of exploration faster compared to sRMC. When the agent learns to solve its target, the curiosity of the agent reduces, and this leads to a reduction of further exploration of the environment. In addition to that, the combined rewards collection by the agent in gRMC is much more higher compared to the sRMC. The objective of  solving the a different environment like gRMC using the same architecture of ICM as used in sRMC is achieved.

\section{Conclusion}
\label{section:conclusion}
RL on industrial scheduling problems are difficult to train because of the inherent bias related to hand tuned rewards. A novel combination of the flexible RMC and a curiosity based RL has been proposed in this study. We found that the application of the curiosity based RL provides good solutions even with very scarce external reward. Additionally, we found that the reward function architecture used in one environment could be easily transferred to another, which was is impossible in hand tuned rewards. In future work, we plan to improve the performance of the agent by using a hybrid reward function providing rewards with little inductive bias for a quicker initial learning but with enough generalized performance. This combined with a gRMC with a custom message passing network could provide highly generalized solutions that can be directly transferable from one robot to another.\par

\printbibliography

@inproceedings{chentanez2005intrinsically,
  title={Intrinsically motivated reinforcement learning},
  author={Chentanez, Nuttapong and Barto, Andrew G and Singh, Satinder P},
  booktitle={Advances in neural information processing systems},
  pages={1281--1288},
  year={2005}
}

@book{sutton2018reinforcement,
  title={Reinforcement learning: An introduction},
  author={Sutton, Richard S and Barto, Andrew G},
  year={2018},
  publisher={MIT press}
}

@inproceedings{burda2018large,
  title={Large-Scale Study of Curiosity-Driven Learning},
  author={Burda, Yuri and Edwards, Harri and Pathak, Deepak and Storkey, Amos and Darrell, Trevor and Efros, Alexei A},
  booktitle={International Conference on Learning Representations},
  year={2018}
}

@inproceedings{bengio2009curriculum,
  title={Curriculum learning},
  author={Bengio, Yoshua and Louradour, J{\'e}r{\^o}me and Collobert, Ronan and Weston, Jason},
  booktitle={Proceedings of the 26th annual international conference on machine learning},
  pages={41--48},
  year={2009}
}

@incollection{Cunha.2020,
 author = {Cunha, Bruno and Madureira, Ana M. and Fonseca, Benjamim and Coelho, Duarte},
 title = {Deep Reinforcement Learning as a Job Shop Scheduling Solver: A Literature Review},
 pages = {350--359},
 volume = {923},
 publisher = {Springer},
 series = {Advances in intelligent systems and computing,   2194-5357},
 year = {2020},
 }

@article{Gabel.2008,
 author = {Gabel, Thomas and Riedmiller, Martin},
 year = {2008},
 title = {Adaptive reactive job-shop scheduling with reinforcement learning agents},
 volume = {24},
 number = {4},
 journal = {International Journal of Information Technology and Intelligent Computing}
}

@INPROCEEDINGS{Gu.2016,
author={S. {Gu} and E. {Holly} and T. {Lillicrap} and S. {Levine}}, booktitle={2017 IEEE International Conference on Robotics and Automation (ICRA)},
title={Deep reinforcement learning for robotic manipulation with asynchronous off-policy updates},
year={2017}, pages={3389-3396}
}

@misc{hameed2020graphs,
      title={Reinforcement Learning on Job Shop Scheduling Problems Using Graph Networks}, 
      author={Mohammed Sharafath Abdul Hameed and Andreas Schwung},
      year={2020},
      eprint={2009.03836},
      archivePrefix={arXiv},
}

@misc{hameed2020gradient,
      title={Gradient Monitored Reinforcement Learning}, 
      author={Mohammed Sharafath Abdul Hameed and Gavneet Singh Chadha and Andreas Schwung and Steven X. Ding},
      year={2020},
      eprint={2005.12108},
      archivePrefix={arXiv},
}

@article{mnih2015human,
  title={Human-level control through deep reinforcement learning},
  author={Mnih, Volodymyr and Kavukcuoglu, Koray and Silver, David and Rusu, Andrei A and Veness, Joel and Bellemare, Marc G and Graves, Alex and Riedmiller, Martin and Fidjeland, Andreas K and Ostrovski, Georg and others},
  journal={nature},
  volume={518},
  number={7540},
  pages={529--533},
  year={2015},
  publisher={Nature Publishing Group}
}

@article{Park.2020,
 author = {Park, In-Beom and Huh, Jaeseok and Kim, Joongkyun and Park, Jonghun},
 year = {2020},
 title = {A Reinforcement Learning Approach to Robust Scheduling of Semiconductor Manufacturing Facilities: IEEE Transactions on Automation Science and Engineering, 1-12},
 pages = {1--12},
 journal = {IEEE Transactions on Automation Science and Engineering},
 }

@inproceedings{Pathak.2017,
 author = {Pathak, Deepak and Agrawal, Pulkit and Efros, Alexei A. and Darrell, Trevor},
 title = {Curiosity-Driven Exploration by Self-Supervised Prediction},
 booktitle = {Proceedings of the IEEE Conference on Computer Vision and Pattern Recognition (CVPR) Workshops},
 year = {2017}
}

@article{schrittwieser2019mastering,
  title={Mastering atari, go, chess and shogi by planning with a learned model},
  author={Schrittwieser, Julian and Antonoglou, Ioannis and Hubert, Thomas and Simonyan, Karen and Sifre, Laurent and Schmitt, Simon and Guez, Arthur and Lockhart, Edward and Hassabis, Demis and Graepel, Thore and others},
  journal={arXiv preprint arXiv:1911.08265},
  year={2019}
}

@INPROCEEDINGS{schwung2017,
  author={D. {Schwung} and F. {Csaplar} and A. {Schwung} and S. X. {Ding}},
  booktitle={2017 IEEE 15th International Conference on Industrial Informatics (INDIN)}, 
  title={An application of reinforcement learning algorithms to industrial multi-robot stations for cooperative handling operation}, 
  year={2017},
  pages={194-199},
}

@inproceedings{schwung2019,
 author = {Schwung, Andreas and Schwung, Dorothea and {Abdul Hameed}, Mohammed Sharafath},
 title = {Cooperative Robot Control in Flexible Manufacturing Cells: Centralized vs. Distributed Approaches},
 pages = {233--238},
 booktitle = {2019 IEEE 17th International Conference on Industrial Informatics (INDIN)},
 year = {2019},
 }

@article{Silver.2018,
 author = {Silver, David and Hubert, Thomas and Schrittwieser, Julian and Antonoglou, Ioannis and Lai, Matthew and Guez, Arthur and Lanctot, Marc and Sifre, Laurent and Kumaran, Dharshan and Graepel, Thore and Lillicrap, Timothy and Simonyan, Karen and Hassabis, Demis},
 year = {2018},
 title = {A general reinforcement learning algorithm that masters chess, shogi, and Go through self-play},
 pages = {1140--1144},
 volume = {362},
 journal = {Science},
}

@inproceedings{Todorov.2012,
 author = {Todorov, Emanuel and Erez, Tom and Tassa, Yuval},
 title = {MuJoCo: A physics engine for model-based control},
 pages = {5026--5033},
 booktitle = {2012 IEEE/RSJ International Conference on Intelligent Robots and Systems},
 year = {2012},
}

@article{Waschneck.2018,
 author = {Waschneck, Bernd and Reichstaller, Andr{\'e} and Belzner, Lenz and Altenm{\"u}ller, Thomas and Bauernhansl, Thomas and Knapp, Alexander and Kyek, Andreas},
 year = {2018},
 title = {Optimization of global production scheduling with deep reinforcement learning},
 pages = {1264--1269},
 volume = {72},
 number = {1},
 journal = {Procedia CIRP}
}

@article{Xia.2020,
 author = {Xia, Kaishu and Sacco, Christopher and Kirkpatrick, Max and Saidy, Clint and Nguyen, Lam and Kircaliali, Anil and Harik, Ramy},
 year = {2020},
 title = {A digital twin to train deep reinforcement learning agent for smart manufacturing plants: Environment, interfaces and intelligence},
 journal = {Journal of Manufacturing Systems},
 }
\end{document}